\pdfoutput=1
\documentclass[10pt,twocolumn,letterpaper]{article}

\usepackage[pagenumbers]{cvpr} 

\usepackage[accsupp]{axessibility}
\usepackage{graphicx}
\usepackage{amsmath}
\usepackage{dblfloatfix} 
\usepackage{amssymb}
\usepackage{booktabs}
\usepackage{color,soul} 

\usepackage{algorithm}
\usepackage{algpseudocode}
\usepackage{amsmath}

\DeclareMathOperator*{\argmin}{arg\,min}
\newcommand{\indep}{\perp \!\!\! \perp}
\usepackage{booktabs}

\usepackage[pagebackref,breaklinks,colorlinks]{hyperref}

\usepackage[capitalize]{cleveref}
\crefname{section}{Sec.}{Secs.}
\Crefname{section}{Section}{Sections}
\Crefname{table}{Table}{Tables}
\crefname{table}{Tab.}{Tabs.}

\def\cvprPaperID{55} 
\def\confName{EarthVision}
\def\confYear{2022}

\begin{document}

\title{Towards assessing agricultural land suitability with causal machine learning}
\author{Georgios Giannarakis$^1$ \and Vasileios Sitokonstantinou$^{1,2}$ \and Roxanne Suzette Lorilla$^1$ \and Charalampos Kontoes$^1$\\
$^1$BEYOND Center, IAASARS, National Observatory of Athens, Greece\\
$^2$Remote Sensing Laboratory, National Technical University of Athens, Greece\\
{\tt\small \{giannarakis, vsito, rslorilla, kontoes\}@noa.gr}
}
\maketitle

\begin{abstract}

Understanding the suitability of agricultural land for applying specific management practices is of great importance for sustainable and resilient agriculture against climate change. Recent developments in the field of causal machine learning enable the estimation of intervention impacts on an outcome of interest, for samples described by a set of observed characteristics. We introduce an extensible data-driven framework that leverages earth observations and frames agricultural land suitability as a geospatial impact assessment problem, where the estimated effects of agricultural practices on agroecosystems serve as a land suitability score and guide decision making. We formulate this as a causal machine learning task and discuss how this approach can be used for agricultural planning in a changing climate. Specifically, we extract the agricultural management practices of ``crop rotation" and ``landscape crop diversity" from crop type maps, account for climate and land use data, and use double machine learning to estimate their heterogeneous effect on Net Primary Productivity (NPP), within the Flanders region of Belgium from 2010 to 2020. We find that the effect of crop rotation was insignificant, while landscape crop diversity had a small negative effect on NPP. Finally, we observe considerable effect heterogeneity in space for both practices and analyze it.

\end{abstract}

\section{Introduction}
\label{sec:intro}

One of the greatest challenges faced by humankind is producing and supplying food to a rapidly growing population in a climate changing world~\cite{niedertscheider2016mapping}. Given the increasing demand for natural resources, the expansion of croplands exerts substantial pressure on natural ecosystems. Ecosystem prosperity can be compromised by complex human-nature dynamics, risking human well-being itself~\cite{fu2019unravelling}. 

The inclusion of complex causal relationships in environmental decision and policy making processes is key for  policy implementation under sustainable management regimes ~\cite{hunermund2019innovation, zheng2020causal}. Towards this direction, recent research utilizes environmental causal analyses to evaluate potential causes driving an observed or hypothesized change in specific target metrics~\cite{kluger2021combining}. In this paper, we discuss causal inference and machine learning in agricultural policy making that aims at increasing productivity to meet the global food requirements, while ensuring ecosystem resilience. 

Predictive Machine Learning (ML) models can be inefficient for agricultural policy making since they are based on correlations and are not inherently capable of addressing causal questions that a policy maker is after~\cite{hunermund2019causal}. Even when causal analyses are employed, they are usually based on localized experiments on few samples and cannot account for the spatial variability caused by changes in climatic conditions, management practices and other factors ~\cite{deines_satellites_2019}. Policies are often horizontally implemented, leading to the lack of spatial targeting among areas with different ecological characteristics ~\cite{Cullen2018-mc}. To mitigate these issues, the Common Agricultural Policy (CAP) of the European Union (EU) proposes the introduction of agri-environmental measures and eco-schemes that are tailored and optimized to the specificities of different regions ~\cite{ec2018post}. 

Every policy measure aims at implementing an agricultural management practice to achieve a certain goal. In order to assess the efficiency of policy measures for agroecological resilience, policy makers need to know whether particular management practices can achieve a given set of objectives \cite{lampkin2021policies}. Specifically, it is required to know if and how practices are effective at influencing key target metrics, such as the supply of ecosystem services or climate change mitigation and adaptation \cite{scarano2017ecosystem}. There are several studies that approach policy evaluation in experimental settings\cite{colen2016economic}. Nevertheless, large-scale experiments on agricultural policies can be challenging and, thus, the only way to assess their impact is through observational data ~\cite{del2019causal}.

Land and crop information derived from satellite images, and environmental data derived from numerical simulations can both capture large areas with high frequency and at high spatial resolutions. Machine learning based causal inference on observed and simulated data is a promising complementary approach to field experiments, as it enables the detailed assessment of the heterogeneous effects of agricultural practices over millions of fields. The exploitation of this kind of data ensures high spatial representation, accounting for regional differences in climatic, soil and crop conditions. Nevertheless, it should be noted that causal inference on observational data is subject to biases from measurement, selection and confounding ~\cite{rubin2005causal, pearl2009causality}. 

The evaluation of the agricultural practices of Crop Rotation - CR (growing different crops across a sequence of growing seasons) and Landscape Crop Diversity - LCD (increasing the number and evenness of crops grown in a landscape) can be used to allocate them in space. Evaluating such practices by their impact on agroecosystem productivity enables implementing resilient farming systems in the light of climate change and high consumer demand ~\cite{folberth2020global, egli2021more}.

\begin{figure*}[h!]
  \centering
    \includegraphics[width=1\linewidth]{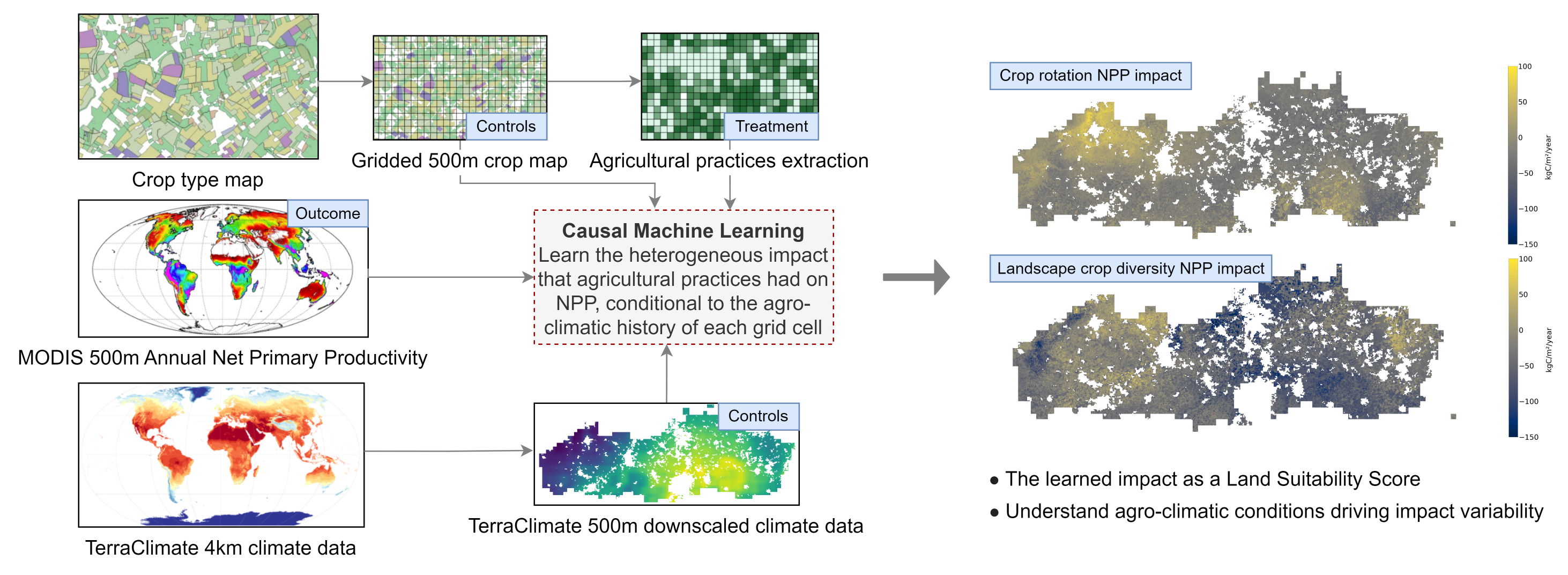}
    \caption{An overview of our process for learning the heterogeneous impact of Crop Rotation (CR) and Landscape Crop Diversity (LCD) on Net Primary Productivity (NPP). The study period extends from 2010 to 2020. First, the parcel-level crop type maps are transformed to gridded maps of 500m cells to match the pixel size of MODIS. For any given year and grid cell, by considering the percentage area of the cell that a crop type occupies, a crop abundance value for all major crop types is generated. Based on this, for each grid cell we compute i) the LCD practice using the Shannon diversity index and ii) the CR practice via quantifying the change between consecutive years in crop abundances within each cell, and summing for the whole study period. Our input data also include environmental variables from the TerraClimate dataset and MODIS NPP values for each cell. We run two separate analyses, one for each practice. Considering the value of each practice as the treatment, NPP as the outcome, and controlling for relevant crop abundances and environmental factors, the dataset is modeled with double machine learning. The heterogeneous impact of a practice per grid cell is studied and used as a land suitability score.}
    \label{fig:method}
\end{figure*}

\noindent \textbf{Contributions.} We demonstrate the applicability of causal machine learning, specifically Double Machine Learning (DML), to estimate the impact that the LCD and CR agricultural practices had on Net Primary Productivity (NPP), which we use as an ecosystem service proxy of climate regulation~\cite{sha2020estimating}. By deriving treatment effects and analyzing their heterogeneity, we infer a data-driven and context-aware agricultural land suitability score for both practices. We thus propose a flexible observation-based agricultural land suitability framework that can be extended to include any management practice and agroecosystem metric that practitioners might have data on. Our approach, illustrated in Figure \ref{fig:method}, was tested in Flanders for the period 2010-2020, where we studied the practices' effect heterogeneity and discussed results in the context of local characteristics.

\section{Related Work}

\noindent \textbf{Agricultural land suitability.} 
Understanding agricultural land suitability is critical for policy applications aiming to foster sustainable agricultural practices and ensure food security. According to Akpoti \etal ~\cite{akpoti2019agricultural}, the analysis of land suitability depends on many factors, including the purpose of the assessment and data availability, and therefore there is no one-size-fits-all approach. Studied techniques are commonly categorized into traditional and modern approaches. Traditional approaches include quantitative and qualitative methods accounting for biophysical characteristics of land ~\cite{danvi2016spatially,el2016mapping}. On the other hand, modern methods exploit a plethora of variables and, according to Mugiyo \etal~\cite{mugiyo2021evaluation}, are categorized into i) computer-assisted overlay mapping, ii) geo-computation or machine learning and iii) Multi-Criteria Decision Making (MCDM). Jayasinghe \etal ~\cite{layomi2019assessment} used the Analytical Hierarchy Process (AHP), which is the most common MCDM method in literature, and the Decision-Making Trail and Evaluation Laboratory (DEMATEL) model to assess the suitability of land for tea crops. Using DEMATEL they were able to visualize interrelationships between factors in a causal diagram. Here, we explicitly frame agricultural land suitability as a causal machine learning task, attempting to spatially optimize a metric of interest via intervening on the land use; and to the best of our knowledge this is the first work that does so.

\noindent \textbf{Heterogeneous Treatment Effects.} The Average Treatment Effect (ATE) has traditionally been the main concern of causal inference. However, it does not necessarily convey all information needed, particularly in cases where treatment effects vary systematically with sample characteristics. This is frequently the case in applications, and treatment effect heterogeneity has been studied in diverse fields such as medicine, social sciences, and economics~\cite{kravitz2004evidence, gabler2009dealing, kent2010assessing, RePEc:nbr:nberwo:12145}. Such methods have also been researched in the context of earth and environmental sciences, where heterogeneity of effects is present. Effect magnitude is determined by various factors, including geography, climate and land cover. For instance, Serra-Burriel \etal~\cite{serra-burriel_estimating_2021} estimated the heterogeneous effects of wildfires and Deines \etal \cite{deines_satellites_2019} found significant variability in the effects of conservation tillage on yield. Both studies used earth observations to uncover the heterogeneity of treatment effects.

\noindent \textbf{Causal Machine Learning.} While statistical methods can be used for the estimation of heterogeneous treatment effects~\cite{crump2008nonparametric, lee2009non, willke2012concepts}, there has been a recent surge of machine learning methods tailored for the same task~\cite{chernozhukov2018double, wager2018estimation, kunzel_metalearners_2019, dudik2014doubly, jesson_quantifying_2021}. Such methods are suitable for high-dimensional data and are able to learn general functional forms, thus taking advantage of the ever-increasing volume of data. Causal machine learning was recently used in environmental studies utilizing both satellite observations and environmental data. The interaction of aerosols and clouds was studied in Jesson \etal~\cite{jesson_using_2021} with causal neural networks~\cite{jesson_quantifying_2021}. In the context of agriculture, the investigation of the effects of conservation tillage on yield in Deines \etal ~\cite{deines_satellites_2019} was carried out with causal forests~\cite{wager_estimation_2018, athey_generalized_2019}. 

By employing remote sensing data and numerical predictions, one can achieve large spatio-temporal coverage that is much needed for effect heterogeneity to manifest. We also took advantage of the scale of earth observations and climate data and combined them with data that reflect historical agricultural use. We were then able to derive the LCD and CR practices, and estimate their heterogeneous impact on the climate regulation for each land sample.

\section{Data \& Methods}\label{3}

We first provide a brief overview of the theory of Conditional Average Treatment Effects in Sec. \ref{3.1}. We then provide details on the data that were used in our approach, as illustrated in Figure \ref{fig:method}. In Sec. \ref{3.2}, we describe the data derived from crop type maps, including i) the LCD and CR agricultural practices that are used as treatments in our analysis and ii) the crop type abundance features that are used as controls. In Sec. \ref{3.3}, we provide information on the NPP, which is the outcome variable in our heterogeneous treatment effects analysis, and in Sec. \ref{3.4} we describe the environmental factors that are also used as controls. The section concludes with Sec. \ref{3.5} and the methodological setup. 

\subsection{Conditional Average Treatment Effects}\label{3.1}

\noindent \textbf{Terminology and Assumptions.} Using the potential outcomes framework~\cite{rubin2005causal}, let $Y(T)$ denote the value (potential outcome) of a random variable $Y$ if we were to treat a unit with a binary treatment $T \in \{0,1\}$. Given a vector of features $X$ describing the units, we want to estimate:

\begin{equation}
    \theta(x) = \mathbb{E}[Y(1) - Y(0) | X = x]
    \label{cate}
\end{equation}
This heterogeneous treatment effect is referred to as the Conditional Average Treatment Effect (CATE)~\cite{abrevaya2015estimating}. There are three important assumptions: 
\begin{table}[h!]
  \centering
  \begin{tabular}{@{}lc@{}}
    \midrule
    Overlap & $0 < \mathbb{P}(T=1|X=x) < 1 \quad \forall x$ \\
    Unconfoudedness & $(Y(1), Y(0))\indep T | X$ \\
    Consistency & $T=t \implies Y = Y(t)$\\
    \bottomrule
  \end{tabular}
  \label{tab:causalassumps}
\end{table}

\noindent Overlap states that for all feature vectors in the population of interest, receiving treatment is possible but not certain. Unconfoudedness assumes that, conditional on $X$, potential outcomes are independent of the treatment. Consistency postulates that, if the treatment is $t$, the observed outcome $Y$ is actually the potential outcome under $t$. Provided these assumptions hold, CATE is identifiable from observational data 
and equals the following statistical estimand~\cite{jesson_using_2021}:
\begin{equation}
    \theta(x) = \mathbb{E}[Y | T =1, X = x] - \mathbb{E}[Y | T = 0, X = x]
\end{equation}
CATE can be similarly defined for continuous treatments, and by averaging over $X$ it reduces to the standard ATE $\mathbb{E}\big[Y(1) - Y(0) \big]$. Variables included in the feature vector $X$ are known as controls. Depending on the causal structure of the phenomenon under study, controlling for a variable might reduce or increase bias of effect estimates~\cite{cinelli_crash_2020}.

\noindent\textbf{Double Machine Learning.} Assuming unconfoudedness, Double Machine Learning~\cite{chernozhukov2018double} formulates the data generating process in terms of the Partially Linear Model~\cite{robinson1988root}:

\begin{align}
    Y &= \theta(X)\cdot T + g(X) + \varepsilon \label{doublemllinearity}\\
    T &= f(X) + \eta \label{confounding}
\end{align}
where $\theta(X)$ is the CATE, and $g,f$ are arbitrary functions (nuisance parameters). Notably, \eqref{confounding} keeps track of confounding as features $X$ drive both the treatment $T$ and outcome $Y$. CATE $\theta(X)$ is then estimated using a two-stage estimation procedure. During the first stage, both the outcome $Y$ and treatment $T$ are separately predicted from features $X$ using arbitrary ML models. Then, in the final stage, the CATE $\theta(X)$ is estimated by solving (for the linear case)

\begin{equation}
    \hat{\theta} = \argmin_{\theta \in \Theta}\mathbb{E}\big[(\Tilde{Y} - \theta(X)\cdot \Tilde{T})^2 \big]
    \label{doublemlfinalstage}
\end{equation}
over a model class $\Theta$, where $\Tilde{Y}$ are the residuals of the $Y \sim X$ regression, and $\Tilde{T}$ are the residuals of the $T \sim X$ regression. In \eqref{doublemllinearity} and \eqref{doublemlfinalstage} the linearity assumption can be dropped to allow for fully non-parametric CATE estimation.

\subsection{Data derived from crop type maps}\label{3.2} \label{subsection:croppractices}

Co-designing practical and effective solutions to major challenges, such as climate change, requires understanding the pattern of interventions carried out by agents of change in agricultural ecosystems, i.e farmers \cite{bohan2022designing}. LCD and CR are considered to be important management practices since they support synergistic improvements in crop yield, environmental health, and ecological sustainability \cite{bowles2020long, tamburini2020agricultural, egli2021more}. Hence, such practices should be considered in strategies that promote sustainable agricultural production. 

We exploited a series of yearly Land Parcel Identification System (LPIS) data to produce variables representing agricultural management practices \cite{mikolajczyk2021species}. LPIS is a geo-spatial database that contains the geometries of the parcels and the declared crop type by the farmers, as part of their  application for CAP subsidies \cite{rousi2020semantically}. 
For each year, we extracted the proportion of the total area occupied by each crop type (crop abundance) per grid cell of the 500m MODIS square grid. We then calculated the Shannon diversity index (H')~\cite{shannon1948mathematical, morris2014choosing} as a metric of LCD per grid cell. We finally took the mean of all years to end up with an average LCD value per cell for 2010-2020. For CR, we summed the absolute difference per crop type abundance per grid cell for two consecutive years; this procedure was repeated for all pairs of adjacent years with which we calculated the total (sum) rotations for the studied period. Both LCD and CR were used as treatments to test their effect on NPP, while crop type abundances for the major crop types in Flanders (grassland, maize, potato, wheat) were used as controls.

\subsection{Net Primary Productivity}\label{3.3}

Net primary productivity (NPP) is the uptake flux, in which carbon from the atmosphere is sequestrated by plants through the balance between photosynthesis and plant respiration. It is a fundamental ecological variable in biosphere functioning, the quantification of which is needed for assessing the carbon balance at regional and global scales\cite{yuan2021effects}. As NPP has been widely used for measuring vegetation dynamics, this index is highly suitable for capturing environmental changes due to natural and anthropogenic factors \cite{gang2017modeling,zhou2021identifying}. The MODIS NPP is produced by the US National Aeronautics and Space Administration (NASA) Earth Observing System. It is based on an energy  budget approach, utilizing earth observations on the fraction of photosynthetically active solar radiation absorbed by the vegetation surface \cite{pan2006improved,running2004continuous}. The results have been validated as being able to capture spatio-temporal patterns across various biomes and climate regimes \cite{zhao2005improvements}. We used annual time series of MODIS NPP (MOD17A3HGF v006) gridded at 500m for 2010-2020 as the target metric (outcome) representing the ecosystem function of climate regulation over Flanders (North Belgium). The MOD17A3 annual product is derived by summing all 8-day net photosynthesis products (MOD17A2H), partly capturing seasonal variability.

\subsection{Environmental factors}\label{3.4}

Variability in agricultural NPP arises due to environmental effects, with specific conditions favoring specific crop types. To uncover the diverse environmental conditions that support or inhibit climate regulation we used a series of parameters provided by Terraclimate\cite{abatzoglou_terraclimate_2018}. The gridded meteorological data  were produced through a climatically aided spatiotemporal interpolation of the WorldClim datasets to estimate monthly time series. Environmental factors used in this study included maximum and minimum temperature, actual evapotranspiration, climate water deficit, precipitation, soil moisture, downward surface shortwave radiation and vapor pressure. The idea behind the selection of the aforementioned variables relies on the wide recognition that temperature and precipitation directly affect NPP \cite{chu2016does, gholkar2014influence, yuan2021effects}, the influence of water availability on soil productivity and vegetation growth \cite{hailu2015reconstructing}, and the processes themselves (e.g. evapotranspiration) that relate to the ecosystem function of climate regulation \cite{yang2020emergy,sun2017ecohydrological}.

\subsection{Methodological setup}\label{3.5}

We have thus created a population of $N$ units indexed by $i$, where $Y_i(T_i)$ is the observed outcome of unit $i$ treated with $T_i$ and $X_i$ are the controls and features that generate systematic variation on the treatment effect. Specifically, the population consists of $N$ grid cells exhaustively covering agricultural land within Flanders, $Y_i$ is the observed NPP value of each cell, $T_i$ is (the value of) the agricultural practice that was applied to the cell, and $X_i$ refers to important characteristics of the specific grid cell, comprising environmental data and agricultural use (i.e., crop types cultivated within the grid cell). The CATE \eqref{cate} is then the average impact that practice $T_i$ had on NPP $Y_i$ for the $i$-th grid cell, conditional to its characteristics $X_i$. Obtaining such local insights on where and why the impact magnitude of agricultural practices on important agro-ecosystem metrics differed is of primary interest to policy makers as it allows for targeted agricultural policy making. 

All data mentioned in the previous subsections are retrieved as time series, with different spatio-temporal resolutions. Data were temporally aggregated over the period of study (2010-2020), since treating every cell-year combination as different units would potentially introduce interference effects~\cite{tchetgen2012causal}, where treatments (agricultural practices) during any year might also influence the outcome (climate regulation) of the next years. For the case of CR, temporal aggregation over the 2010-2020 period happens by summing all the rotations that happened in grid cell $i$ from 2010 to 2020, while for LCD we consider the average Shannon diversity index over all years. This is the treatment $T$ we are using for the analyses.

By adding the crop abundances to the feature vector $X$, we control for the four most dominant crop types mentioned in Sec. \ref{3.2}; the median abundance value of all other crop types is less than $2\%$. We also control for important environmental variables, listed in Sec. \ref{3.4}. Crop types are confounders for the causal relation under study and thus good controls as they drive the magnitude of both the agricultural practices and the NPP. As such, they should be included in the model. On the other hand, environmental data are not driving any practice, but they are driving the NPP values, thus making them strong candidates for good controls \cite{cinelli_crash_2020}. We finally binarize the treatment by letting the median CR value and median LCD value to be thresholds, over which we designate cells as treated units and the rest as control.

The assumption of unconfoudedness can't be tested, and within any observational study it is likely to be invalid. Even if the variables we controlled for will help with bias reduction, some bias from unobserved confounders might still be present. Nevertheless, we note that by being cautious with the selection of controls we try to avoid selection bias, and reported results provide a large-scale complementary approach to localized field experiments.

\section{Experimental Results \& Discussion}

\noindent\textbf{Filtering.} For the experiments we used the EconML Python package implementing the Double Machine Learning method \cite{econml}. We derived CATE estimates for both LCD and CR at the 500m native NPP resolution. 
We ensured that fitting takes place over actual croplands by restricting the dataset to grid cells where the sum of crop abundances, given by the crop type maps, exceeded a threshold of $80\%$. To aid the overlap assumption, propensity scores were first estimated using a Gradient Boosting Propensity Model from the CausalML Python package~\cite{chen2020causalml}, and units with extreme propensity scores ($\leq 0.2$ or $\geq 0.8$) were filtered out. Within DML, it is crucial to avoid overfitting first stage models; otherwise a portion of the outcome and treatment variability explained will be due to factors other than the controls. During the first stage of DML only, we therefore split the dataset to train and test (80-20) to evaluate the predictive performance and assess overfit.

\noindent\textbf{Double Machine Learning.} In first stage, for both the CR and LCD analyses, random forest regressors were selected to predict NPP from controls, outperforming Lasso and gradient boosting regression. To predict the binary treatments themselves, logistic regression was used, outperforming random forest and gradient boosting classifiers.

All model selection procedures happened by performing 3-fold cross validation and a grid search for hyperparameter optimization. During the first stage only, the maximum absolute scaler was applied, as it retains the sparsity structure that is prevalent in crop abundance data. For the final stage regression, for both analyses, a Causal Forest \cite{wager_estimation_2018} with $1000$ trees was used with heterogeneity score as the splitting criterion. The final stage causal forest was fine-tuned based on the out of sample performance. Minimization of \eqref{doublemlfinalstage} happened with unscaled features to maintain interpretation in the original units of measurement. Fitting results for CR and LCD  can be found in Table~\ref{tab:dmlfits}. First stage models captured a significant part of the variability of both the outcome and treatment variables in both cases. The difference between the training and test performance was marginal, indicating that models avoided overfitting.

\begin{table}[h!]
  \centering
  \begin{tabular}{@{}lcr@{}}
    \toprule
   \textbf{Crop Rotation} & Train & Test\\
    \midrule
    $Y \sim X$ (Outcome Modeling) & $0.75$ & $0.74$\\
    $T \sim X$ (Treatment Modeling) &  $0.63$ & $0.62$ \\
    \toprule
    \textbf{Landscape crop diversity}  & Train & Test\\
    \midrule
    $Y \sim X$ (Outcome Modeling) & $0.73$ & $0.71$\\
    $T \sim X$ (Treatment Modeling) &  $0.59$ & $0.62$ \\
    \bottomrule
  \end{tabular}
  \caption{Performance ($R^2$ for outcome modeling, F-1 score for treatment modeling) of ML models internally used by DML. Outcome $Y$ is Net Primary Productivity, treatment $T$ is crop rotation / landscape crop diversity, $X$ is a vector of features.}
  \label{tab:dmlfits}
\end{table}

\begin{figure*}[!b]
  \centering
  \begin{subfigure}{0.8\linewidth}
    \includegraphics[width=1\linewidth]{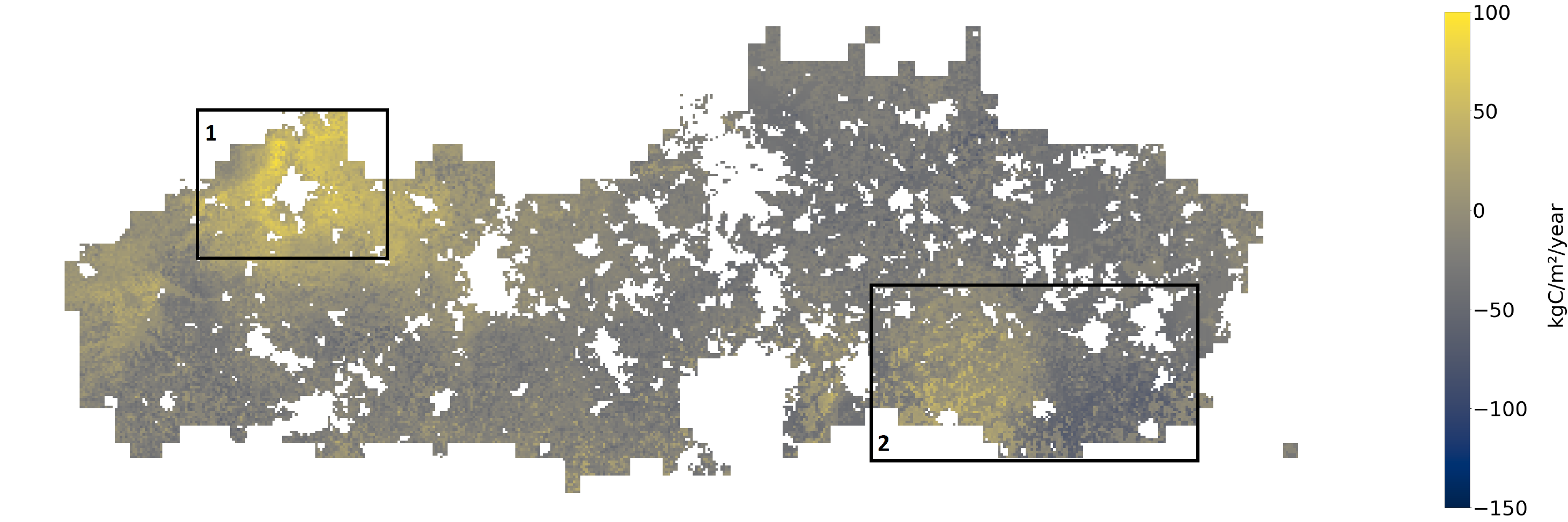}
    \caption{Impact of crop rotation on climate regulation (NPP) in Flanders at 500 m resolution.}
    \label{fig:map-a}
  \end{subfigure}
  \hfill
  \begin{subfigure}{0.8\linewidth}
    \includegraphics[width=1\linewidth]{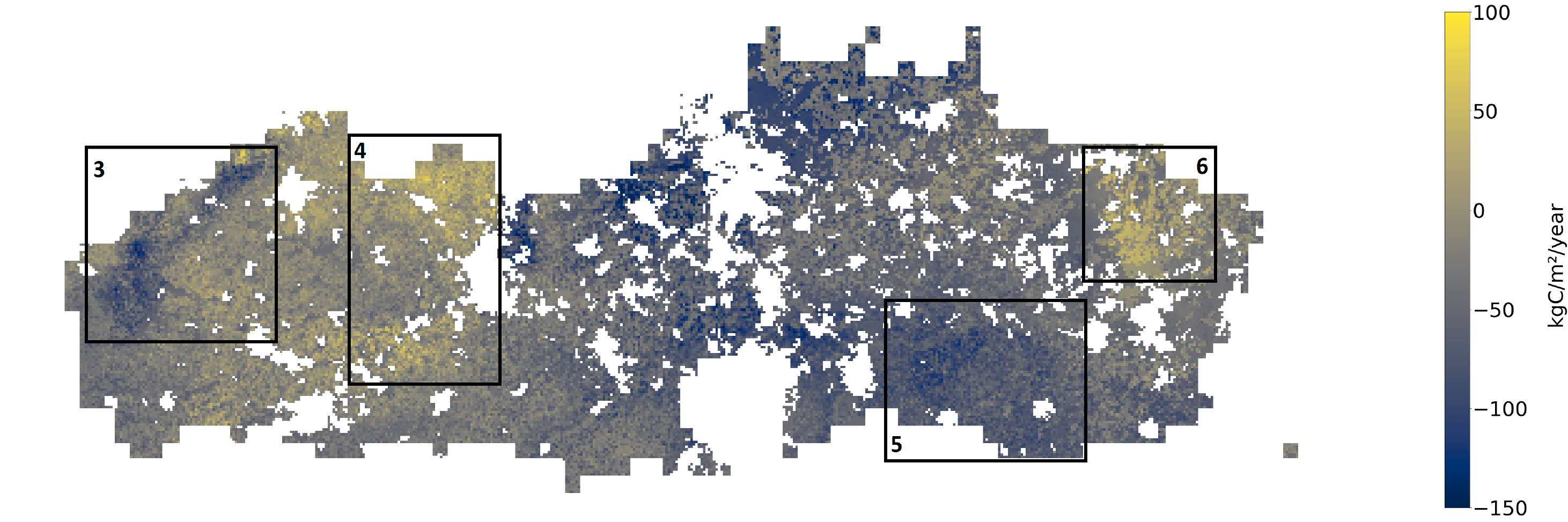}
    \caption{Impact of landscape crop diversity  on climate regulation (NPP) in Flanders at 500 m resolution.}
    \label{fig:map-b}
  \end{subfigure}
  \caption{Estimated impact of crop rotation (a) and landscape crop diversity (b) on climate regulation (NPP) based on the heterogeneous treatment effect analysis for the Flanders region from 2010 to 2020. The annual NPP outcome variable is measured in kgC/m²/year.}
  \label{fig:map}
\end{figure*}

\subsection{CATE Results}\label{4.1}

Figure~\ref{fig:map} illustrates the DML CATE estimates over the region of Flanders for LCD and CR. Intervening by applying a practice in a cell with a CATE estimate of e.g. $20$, we would expect an increase of $20$ kilograms of carbon for each squared meter of land within a year. From such results, a spatially explicit policy for sustainable agriculture can be extracted, by prioritizing for each cell the application of practices whose effect estimate (land suitability score) is high. Such analysis can be repeated for any agricultural practice and outcome metric of interest, while appropriate controls can be added using expert knowledge.

Our results indicated that in north-west areas that are characterised by high abundance of grasslands, CR increased NPP by approximately 100 kgC/m²/year (Figure \ref{fig:map-a}; square 1). This suggests that the contribution of grasslands in crop rotations is important in positively driving climate regulation \cite{zarei2021evaluating}. Positive effects were also found on south-east areas where the dominating crop type is winter wheat followed by maize and grasslands (Figure \ref{fig:map-a}; square 2). 

For LCD, negatively affected areas west (Figure \ref{fig:map-b}; square 3) and south-east (Figure \ref{fig:map-b}; square 5) that are dominated by grasslands and winter wheat revealed a decrease in NPP of over 100 kgC/m²/year. Due to the importance of the aforementioned crop types to the carbon cycle, one could expect positive LCD impacts on local climate regulation \cite{bengtsson_grasslandsmore_2019, schmidt2012carbon}. However, as these crops appear clustered in space, diversifying such cropland regions led to a negatively affected NPP.
By contrast, where multiple crop types exist, maize showed a significant role in carbon uptake as its existence in a diversified agricultural landscape indicated an NPP increase of approximately 50 kgC/m²/year (Figure \ref{fig:map-b}; square 4 and 6). In fact, maize has a high capacity in capturing large amounts of carbon from the atmosphere \cite{lal2010managing}.

\begin{figure}[h!]
  \centering
  \includegraphics[width=1\linewidth]{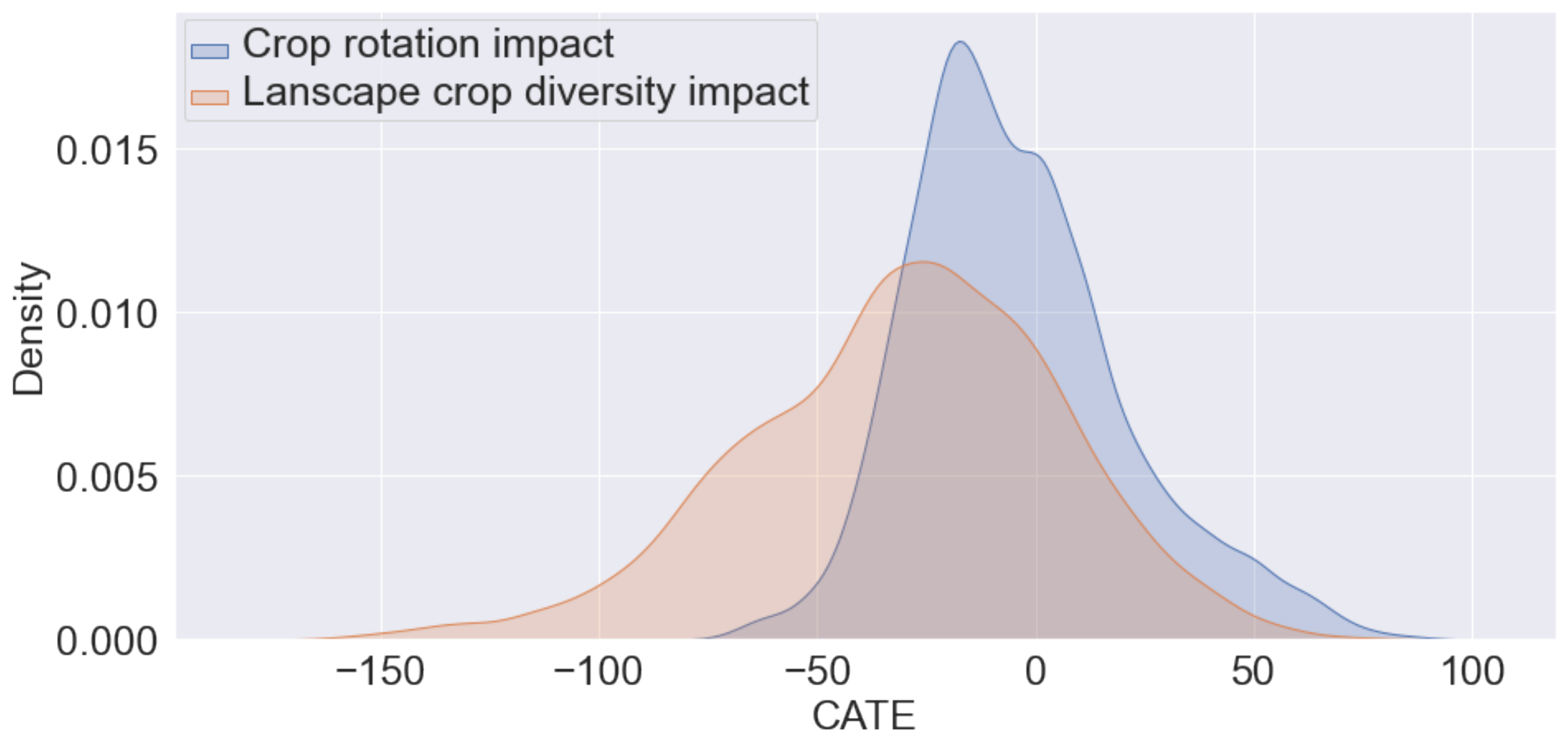}
    \caption{Distribution of CATE estimates for crop rotation and landscape crop diversity.}
  \label{fig:rota500distr}
\end{figure}
\begin{figure*}[b!]
  \centering
  \begin{subfigure}{1\linewidth}
    \includegraphics[width=1\linewidth]{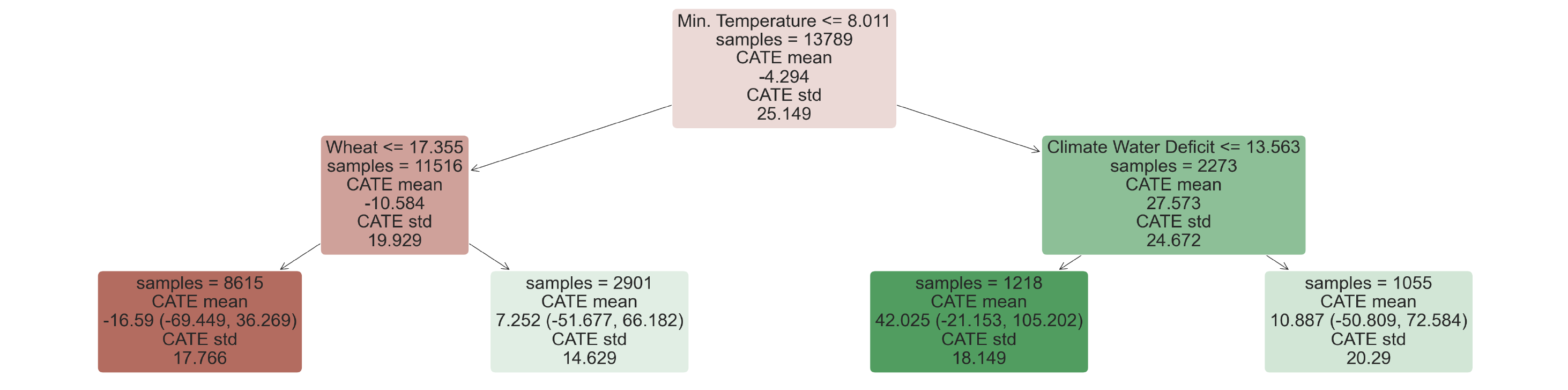}
    \caption{Heterogeneity tree for crop rotation.}
    \label{fig:tree-a}
  \end{subfigure}
  \hfill
  \begin{subfigure}{1\linewidth}
    \includegraphics[width=1\linewidth]{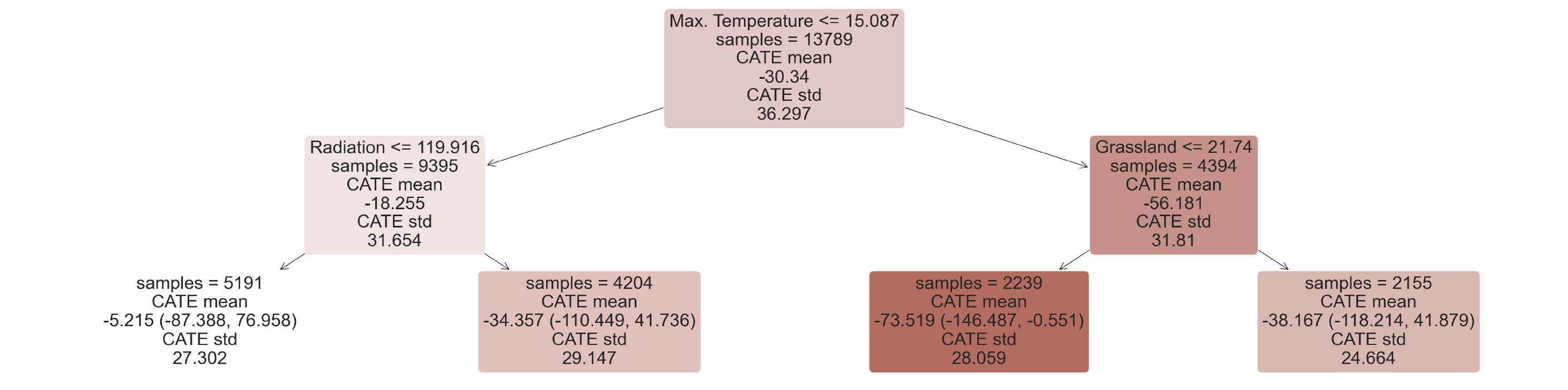}
    \caption{Heterogeneity tree for landscape crop diversity.}
    \label{fig:tree-b}
  \end{subfigure}
  \caption{Trees interpreting effect heterogeneity for both practices studied. To be read from top to bottom, going left if the Boolean condition at the top of each box is true, and right if it is false. Sample size for each leaf is reported, alongside the CATE mean, standard deviation and confidence intervals. Wheat and grassland abundance are in \%, min. and max. temperature are in °C, climate water deficit is in mm and radiation is in W/m².}
  \label{fig:tree}
\end{figure*}
The ATE estimates for CR were $1.08$ $(95\% \text{ CI } [-20.35, 22.51])$ and for LCD were $-35.73$ $(95\% \text{ CI } [-58.73, -12.73])$. The distribution of CATE estimates for both practices over the Flanders croplands can be seen in Figure \ref{fig:rota500distr}.
Besides the more granular CATE insights we are reporting, we also note that a LCD practice implemented over the entire Flanders cropland would have a weak yet significant negative impact on NPP. On the other hand, a CR practice implemented across the entire terrain would be able to bring multiple agro-environmental benefits without impacting the average NPP \cite{karlen2006crop, havlin1990crop}.

\subsection{Heterogeneity Analysis}

We used causal trees to analyze the heterogeneity of the model, accounting for all features in $X$ ~\cite{athey_recursive_2016, econml}. The tree successively splits on feature values that maximize the treatment effect difference across leaves. We constrained the tree depth to $2$ to retain explainability. 

In Figure~\ref{fig:tree-a} we see that minimum temperature was found to be the most important driver of effect heterogeneity for CR. Particularly, high minimum temperatures combined with low climate water deficit seem to benefit CR performance in Flanders. Wheat abundance was also detected as a major contributor to CR heterogeneity, spatially coinciding with high CATE estimates as reported in Sec. \ref{4.1}. 

LCD impact heterogeneity (Figure~\ref{fig:tree-b}) was found to be affected by maximum temperature, solar radiation and grassland abundance. LCD negatively affected NPP the least when maximum temperature was lower than 15.1\textcelsius, and radiation was lower than 119.9 W/$m^2$. Additionally, in high maximum temperatures, diversifying crops in grassland-abundant cells appears more preferable than diversifying cells lacking grasslands.

\begin{figure}[h!]
  \centering
  \includegraphics[width=1\linewidth]{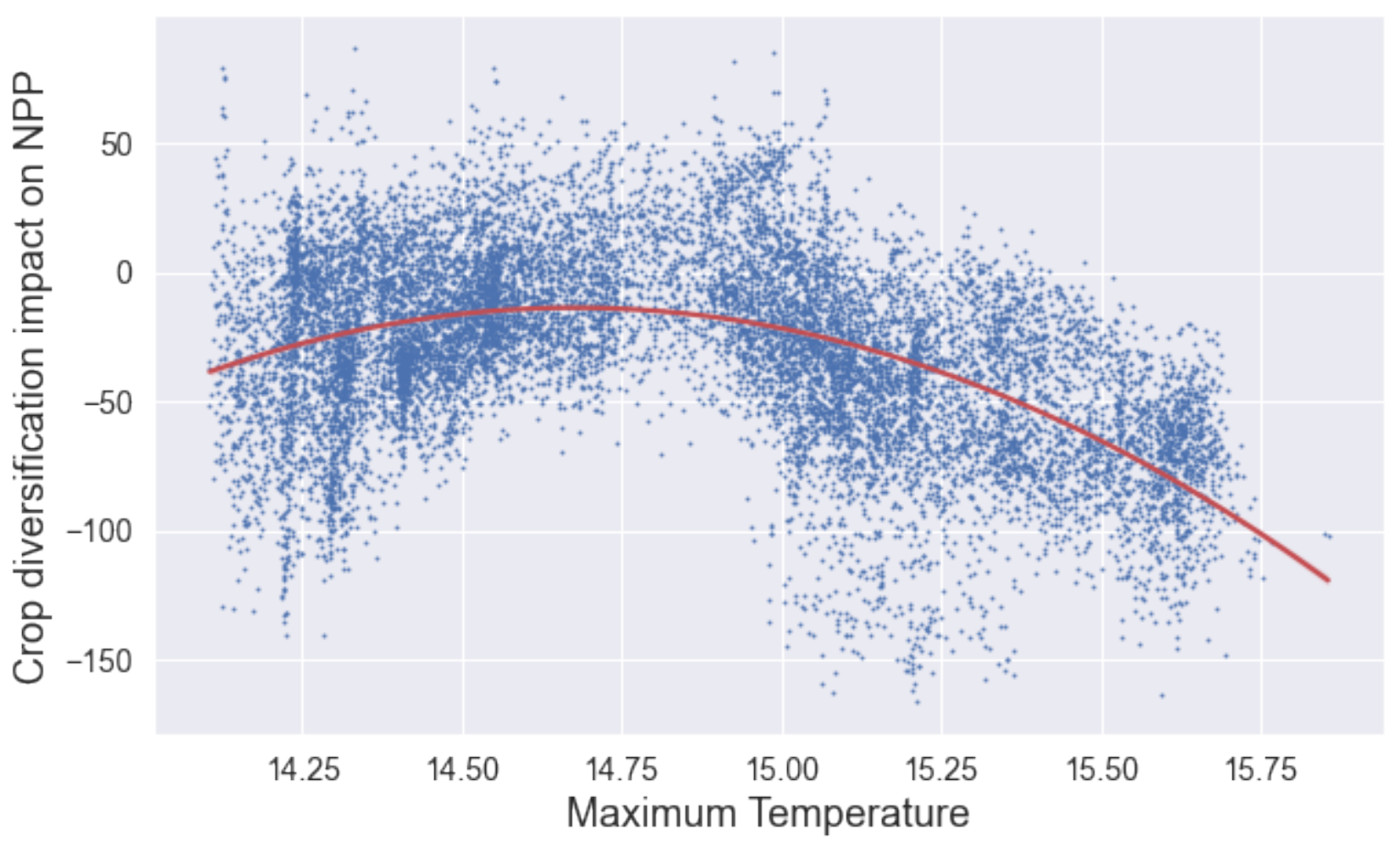}
    \caption{Landscape crop diversity impact against max. temperature (°C).}
  \label{fig:div500tmax}
\end{figure}

\begin{figure}[h!]
  \centering
  \includegraphics[width=0.90\linewidth]{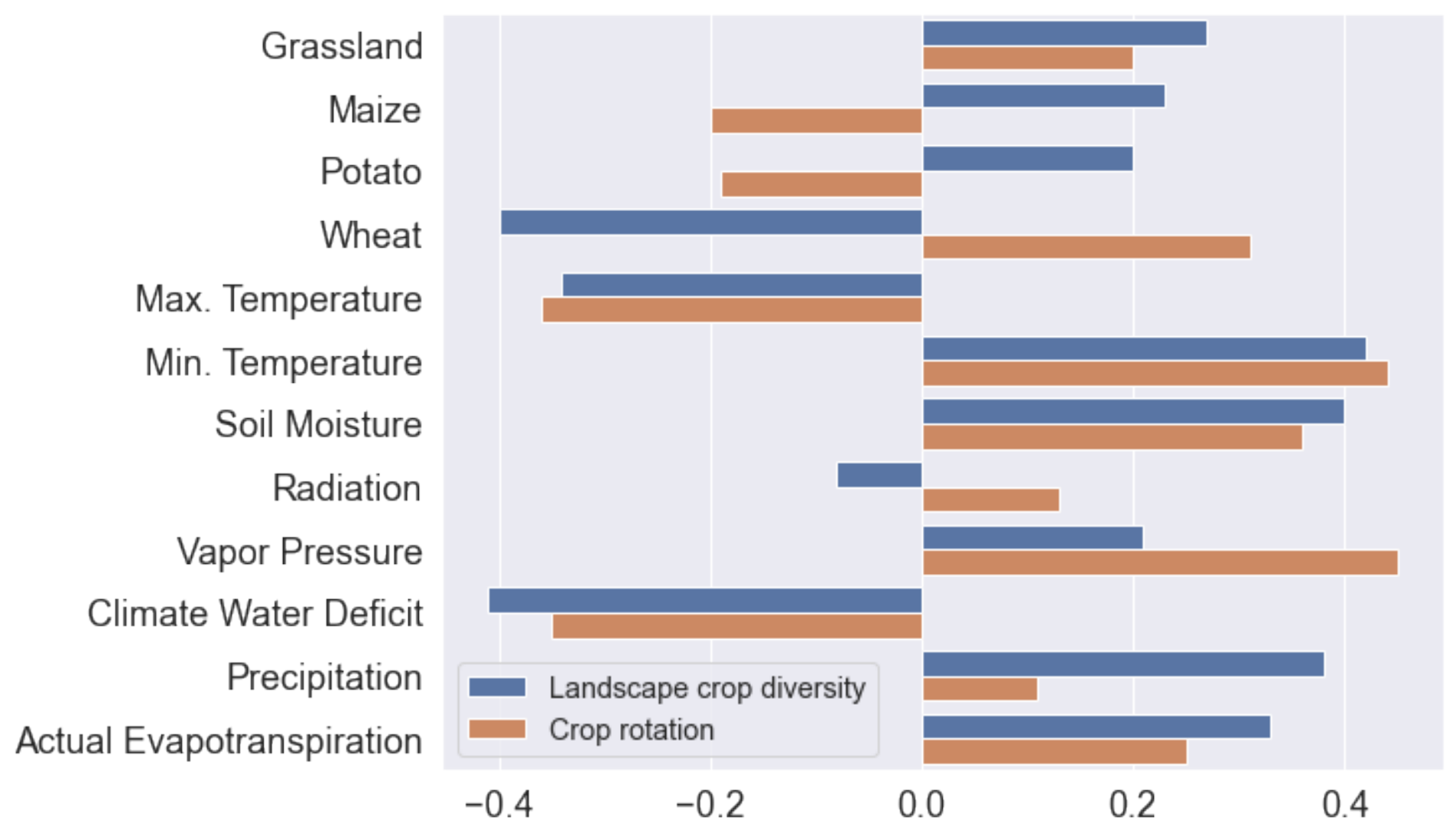}
    \caption{Barplot showing Spearman correlation coefficients of estimated CATEs with all features for both practices. }
  \label{fig:barplot}
\end{figure}

While trees are useful to detect the most important features in the sense described above, a complementary understanding of effect heterogeneity can be obtained by plotting CATE estimates as a function of selected features. In Figures \ref{fig:rota500tmax} and \ref{fig:div500tmax}, we see clear patterns that show maximum temperature driving effect heterogeneity for both practices. For the rest of the features, we report Spearman correlation coefficients between them and the estimated CATE in Figure \ref{fig:barplot}. There are moderate correlations for almost all factors, indicating that the effect under study is indeed heterogeneous. The correlation values for environmental features are larger but comparable to the ones of crop features, highlighting that both control categories contribute to heterogeneity.

\subsection{Agricultural land suitability \& climate change}

So far, we estimated the impact of agricultural practices on climate regulation from historical data (2010-2020). 
Nevertheless, important climatic variables are expected to change over the next decades as a result of climate change. The estimation of their trajectory has been the subject of numerous scientific studies \cite{stocker2014climate}. In order to devise impactful agricultural policies, we need to evaluate the performance of practices in future climatic conditions.

In the context of causal machine learning, we learned the function $\theta(x) = \mathbb{E}[Y(1) - Y(0) | X = x]$ where feature vector $X$ contains multiple climate variables. In theory, we are able to derive the impact of a practice on an outcome metric under future climatic conditions by changing the relevant variables of feature vector $X$ and re-calculating. However, $\theta(x)$ was learned from historical data, and by definition future climatic conditions were not observed. By changing climate variables to match future projections, we are essentially extrapolating the function to points outside the observed feature space. 

\begin{figure}[h!]
  \centering
  \includegraphics[width=1\linewidth]{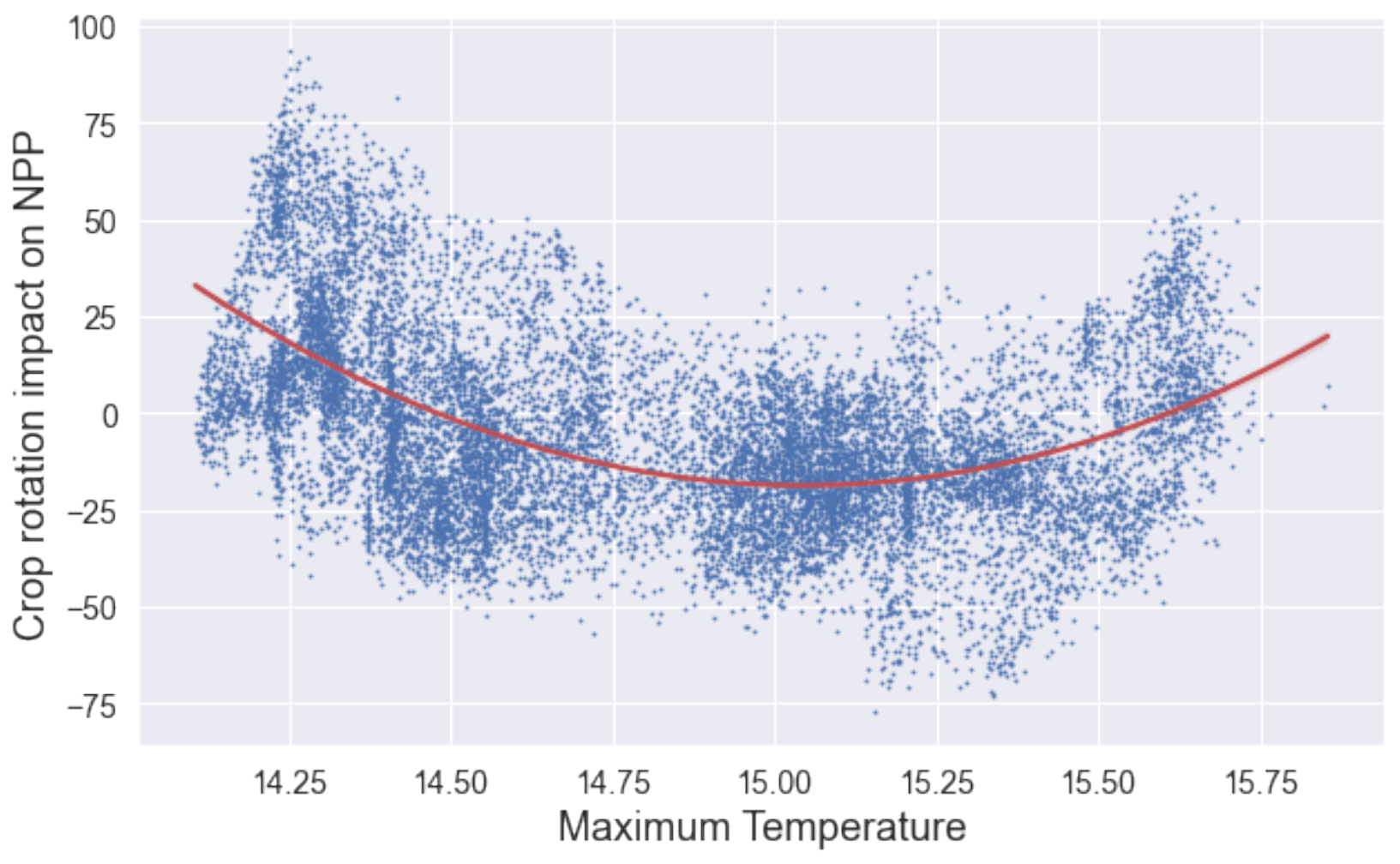}
    \caption{Crop rotation impact against max. temperature (°C).}
  \label{fig:rota500tmax}
\end{figure}

In Figure \ref{fig:rota500tmax} for example, a quadratic trend between the CR CATE and maximum temperature is seen, allowing us to hypothesize that in slightly warmer conditions the benefit of CR would increase. Thus, our results provide preliminary insights on an ex-ante impact assessment for the agricultural practices studied in the context of climate change, enabling the formulation of scientific hypotheses for further study. 

Future work includes incorporating climatic projections in the pipeline and comparing the impact learned from them to the impact learned from historical data. Further work also comprises the study of more agricultural practices, e.g., intercropping and minimum soil cover, but also the study of more target outcomes, such as erosion prevention and pollination potential. This way, we can move towards a synthesis of ecosystem service trade-offs and the formulation of a comprehensive suite of policies in response to climate change. Finally, the correlations between agricultural practices and NPP, and consequently the treatment impact, depend on the spatial scale of analysis \cite{zhou2021spatial}. Therefore, agricultural practices should be evaluated at different scales to gain insights towards optimizing agriculture land suitability from farm to landscape level and achieve optimum value.

\section{Conclusions}

We presented an approach for assessing agricultural land suitability using causal machine learning.  We estimated the heterogeneous treatment effects of CR and LCD on climate regulation (NPP), accounting for historical crop and environmental data. We used earth observations (MODIS), climate data (TerraClimate) and openly available yearly crop type maps covering North Belgium (Flanders) from 2010 to 2020. CATE results reveal significant impact heterogeneity in space, highlighting the usefulness of extended spatio-temporal data coverage, the importance of spatially targeted measures and the relevance of CATE estimates as land suitability scores.
The ATE for CR is insignificant, while for LCD it is significant but relatively small compared to the influence environmental drivers have on NPP \cite{zhou2021identifying}. It would be difficult to extract such subtle effects from localized experiments of confined samples.

Significant challenges remain towards the application of the proposed approach in real-life planning. Observational studies enable the manifestation of heterogeneity, which is vital for spatially explicit recommendations, but suffer from unobserved confounding. For this reason, our results should be used along with insights from local field experiments. Furthermore, we offered insights on how our two treatments affect climate regulation but we did not provide detailed information on the contribution of specific crop types. For the CR treatment, we do not consider the crop types involved in the transition but simply the transition itself. Similarly, for the LCD all crop type mixtures are treated equally. Additional analyses with crop-aware treatments can offer enhanced understanding for evidence-based decision making.

\section*{Acknowledgements}

This work has been supported by the EIFFEL project (EU Horizon 2020 - GA No. 101003518) and the MICROSERVICES project (2019-2020 BiodivERsA joint call, under the BiodivClim ERA-Net COFUND programme, and with the GSRI, Greece - No. T12ERA5-00075).

{\small
\bibliographystyle{ieee_fullname}
\bibliography{references}
}

\end{document}